\pdfoutput=1

\documentclass[11pt]{article}

\usepackage[]{acl}

\usepackage{placeins}
\usepackage{times}
\usepackage{latexsym}
\usepackage{amssymb}
\usepackage{amsmath}
\usepackage{caption}
\usepackage{subcaption}
\usepackage{booktabs}
\usepackage[T1]{fontenc}

\usepackage[utf8]{inputenc}

\usepackage{microtype}

\usepackage{graphicx}
\newcommand{\cready}[1]{{}}
\graphicspath{{graphs/}}
\title{Fusing finetuned models for better pretraining}

\author{\bf{Leshem Choshen\thanks{\ \ These authors contributed equally to this work.}, Elad Venezian\footnotemark[1], Noam Slonim, Yoav Katz} \\
\\
IBM Research \\
\{leshem.choshen, eladv, katz, noams\}@il.ibm.com}

\begin{document}
\maketitle
\begin{abstract}
Pretrained models are the standard starting point for training. This approach consistently outperforms the use of a random initialization. However, pretraining is a costly endeavor that few can undertake.

In this paper, we create better base models at hardly any cost, by fusing multiple existing fine tuned models into one. Specifically, we fuse by averaging the weights of these models. We show that the fused model results surpass the pretrained model ones. We also show that fusing is often better than intertraining.

We find that fusing is less dependant on the target task. Furthermore, weight decay nullifies intertraining effects but not those of fusing.





\end{abstract}

\section{Introduction}

Over the past decade, pretrained models has dominated the world of NLP. Such pretrained models are first trained on a \emph{source task}. Then, they serve as an initialization point for training on a new \emph{target task}. Training the \emph{base model} parameters on a target task is termed \emph{finetuning} and is the current most effective way of training \citep{Chen2022RevisitingPT}, yielding higher performance and more stable training process, with less labeled data. 

In this work, we present a novel approach for weight initialization, leveraging the preexisting finetuned models. Specifically, we propose in Section~\S\ref{sec:fusing} to fuse together many models finetuned on source tasks to leverage the resources invested in their training process, for future target tasks. While pretrained models are few, and pre-training new models is expensive, existing finetuned models are abundant (App.~\ref{sec:tuning_freq}) and can be easily reused.

\begin{figure}[tbp]
\includegraphics[width=\columnwidth]{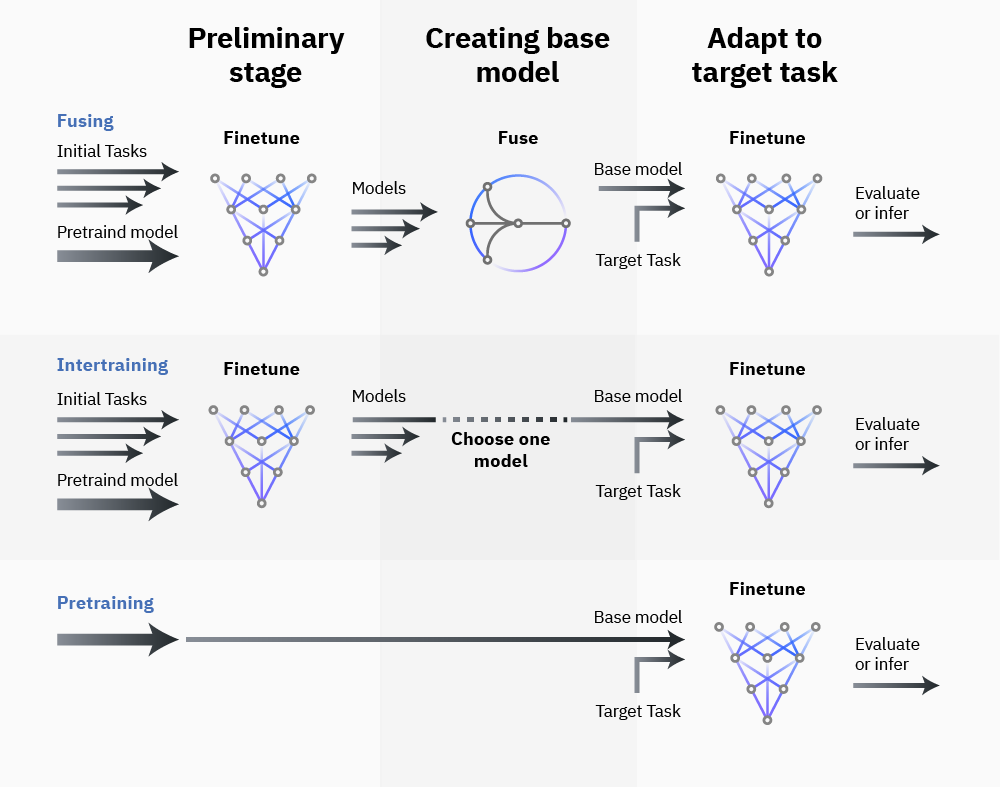}
\caption{Different uses for available finetuned models. The simplest starts from a \textit{pretrained} model ignoring the finetuned models (bottom), \textit{intertraining} picks one model (center), \textit{Fusing} takes the finetuned models and combines them (top). Then they all use the model as a base model for finetuning on the target task.
\label{fig:fusing_sheme}}
\end{figure}

In a way, our work, reverses the transfer learning paradigm. Instead of reusing a pretrained model to create better finetuned models, we propose reusing the finetuned models to create a better pretrained model.

A well known approach for selecting a weight initialization, is called intertraining \citep{Phang2018SentenceEO}. In this approach, the base model is a fine tuned model of another task, preferably related to the target task. Fusing generalizes intertraining (See Fig.~\ref{fig:fusing_sheme}). Fusing is combining several finetuned models, while intertraining can be seen as special case of "combining" only a single model.

We experiment with fusing on 3 families of datasets consisting of 30 datasets (\S\ref{sec:results}). We find that fusing consistently outperforms pretraining in both accuracy and stability on T5. Further, we find fusing to be often better than intertraining, when fused models are chosen carefully. Specifically, fusing two models, is often better than intertraining with either one of them. Fusing the best two finetuned models for intertraining led to the best results in our experiment.

Last, we find that our use of weight decay in finetuning hurts intertraining considerably (\S\ref{sec:decay}), while fusing seems robust to this.

\section{Fusing}\label{sec:fusing}



We define model Fusion (See Fig.~\ref{fig:fusing_sheme}) as the process of taking several finetuned models and creating a new base model. Ideally, the fused model would be a better base model than any single finetuned model it was based on and better than the original pretrained model.

Formally, given an initialization base model $P$ and $n$ models finetuned on it, let $W_1,W_2\ldots W_n\in\mathbb{R}^d$ be the weights finetuned by the models over $P$. Fusing is a function 
\begin{align*}
    W_{fuse} &= f(W_1, W_2, \ldots , W_n) \\
    &\mathbb{R}^d \times \mathbb{R}^d \times \ldots \times \mathbb{R}^d  \rightarrow \mathbb{R}^d
\end{align*}

In this work, we propose the simplest form of fusion. For each weight shared by all models, assign the average weight to the model. 
\begin{align*}
    W_{fuse} &=  f\left(W_1,W_2, \ldots , W_n\right) \\
    &=  \frac{W_1 + W_2 + \ldots + W_n}{n}
\end{align*}

\section{Experimental Setup}\label{sec:Exp}

Our main experiments assess which is the best base model for finetuning on a target task. Each experimental setting considers a family of target tasks (\S\ref{sec:tasks}) and a set of available finetuned models (\S\ref{sec:available_models}). We compare fusing to its alternatives (c.f. \S\ref{sec:setup_models}): starting from a pretrained model, or starting from a model finetuned on another task (intertraining).

We finetune each base model on each target task. We report accuracy and often its standard deviation, as both performance and stability are important aspects of a model \citep{Zhang2021RevisitingFB}.

\subsection{Dataset Families} \label{sec:tasks}
Our experiments are based on 3 families of English text classification datasets. We focus on text classification for ease of evaluation, but assume the tasks are diverse enough for conclusions to extend to other settings. 

It is debated whether pretraining on a similar task benefits the target task \citep{krishna2021does,rothe2021thorough,zhang2020pegasus,ram2021splinter}. ***\citet{aribandi2021ext5} showed that families of related tasks correlate with how they improve under different multi-task pretraining scenarios. 
It is hence important to see what kinds of tasks are useful for fusing and whether fusing models with specific traits are more beneficial.

\paragraph{General.} This dataset family contains GLUE \citep{wang-etal-2018-glue} and SuperGLUE text classification datasets \citep{Wang2019SuperGLUEAS} (full list of datasets in App.~\ref{ap:sec:datasets}), excluding test only and regression datasets. The datasets consist of a wide range of classification tasks, from sentiment analysis to linguistic acceptability to natural language inference. It is the most commonly used benchmark in previous works.

\paragraph{NLI.} This dataset family is composed of a set Natural Language Inference and Entailment tasks. This represents a family of the same \emph{task} (Full list in App.~\ref{ap:sec:datasets}). 

\paragraph{Twitter.} This dataset family contains all 11 Twitter datasets collected by TweetEval \citep{barbieri-etal-2020-tweeteval} (Full list in App.~\S\ref{ap:sec:datasets}).  The task range from irony detection to emoji prediction. This represents a family of the same \emph{domain} (Full list in App.~\ref{ap:sec:datasets}).

\subsection{Available finetuned models} \label{sec:available_models}
Each experiment simulates a world where certain finetuned models are already available. In most cases, we simulate these preexisting models by finetuning a pretrained model on each source task, usually the general datasets family. The target task is excluded for the source tasks. 


\subsection{Base Models}\label{sec:setup_models}
We consider 3 base models.

\paragraph{Fused model.}
This model is based on fusing the available finetuned models as described in Section~\S\ref{sec:fusing}.  Unless stated otherwise, we fuse all the available fine tuned models. Note that this subset does not include a model finetuned on the target task by design.  

\paragraph{Pretrained model.}
This baseline takes the pretrained model as the base model. This baseline is far from trivial as it is not clear that combining models in our suggested way will be beneficial at all.  Moreover unless the intermediate task is carefully chosen, this baseline is often better than the intertrained model.

\paragraph{Intertrained model.}
A second baseline selecting one available fine tuned model as a base model.  Which model serves best a given target task is a separate research question \cite{Chang2021RethinkingWI}. Often, when the wrong task is chosen intertraining hurts results\cready{cite shachar's paper with the parser that hurt results}. We chose the heuristic of taking the model finetuned on the largest training set.  We note that the largest dataset in the general dataset family, MNLI, was shown to be the best intertraining task for this set \citep{pruksachatkun2020intermediate}.  The second largest is sst2 and it is used for intertraining when evaluating on MNLI as a target task. Similarly mutatis mutandis  it is ESNLI and MNLI for the NLI group and Sentiment Analysis and Emoji for Twitter.

\subsection{Hyperparameters} \label{sec:setup}
We mainly use as a pretrained model T5v1.1-small following the training hyperparameters from the original paper \citep{Abnar2021ExploringTL}. It is both considered to be better than T5 and was not trained on tasks other than the pretraining task, which would interfere with our experiments. We applied early stopping, evaluated each 50 batches with patience of 50 steps and minimum improvement $0.001$ and used ADAMW \citep{loshchilov2017decoupled} without weight decay and in \S\ref{sec:decay} decay of $0.01$ . 

We repeat most experiments with 5 random seeds and in those cases report the standard deviation.

\section{Results} \label{sec:results}
Our results show that fusing is consistently better than pretraining. Moreover, done well, fusing results are even better than intertraining. Last, fusing is shown to be more robust than intertraining, working well even without careful choice of models or parameters.

\begin{table}
\resizebox{\columnwidth}{!}{
\begin{tabular}{ c c c c c | c }
                &               & General             & NLI            & Twitter            & Average \\
\hline
                & General      & \textbf{72.76}       & 70.29          & 56.39              & \textbf{66.48} \\
Intertrain      & NLI          & 71.18                & \textbf{71.11} & 56.36              & 66.22 \\
                & Twitter      & 69.67                & 67.38          & \textbf{57.18}     & 64.74 \\
\hline
                & General      & 68.12                & 67.95          & 54.71              & 63.59 \\
Fuse            & NLI          & 68.96                & 70.65          & 54.54              & 64.72 \\
                & Twitter      & 64.17                & 66.74          & 52.86              & 61.26 \\
\hline
Pretrain        &              & 63.81                & 67.66          & 55.73              & 62.40 \\
\end{tabular}
}
\caption{The best models to fuse are consistent, the best for intertraining is the most similar per domain. The average accuracy over the target dataset family (columns) given available models finetuned on another family (rows).  Results averaged over 5 runs with different seeds (std reported in Appendix~\S\ref{ap:sec:std}) \label{tab:cross_no_decay}}
\end{table}

We first compare fusing all available models to the baselines (See Table~\ref{tab:cross_no_decay}). We experiment on 3 dataset families (See~\ref{sec:tasks}) the general dataset family, datasets of the same task (Natural language inference) and dataset of the same domain (Twitter). We test all combinations of the target and source dataset families and the corresponding available models (See~\S\ref{sec:available_models}). We remind that the target task is never taken as a source task for our experiments. 

Evidently, such fusing is better than pretraining. However, carefully chosen intertraining is better yet. We note that this is not true for any intertraining model, and many models result in worse than pretraining results \citep{pruksachatkun2020intermediate}. Interestingly, intertraining is sensitive to the target task while fusing is not. The best source task for intertraining is the one similar to the target, while for fusing, a good set of models (NLI) is consistently good.

Next, we consider whether choosing a pair of available models for fusing could improve over intertraining. To reduce the number of experiments we use only the general datasets that come from GLUE.

\begin{figure}[tbp]
\includegraphics[width=\columnwidth]{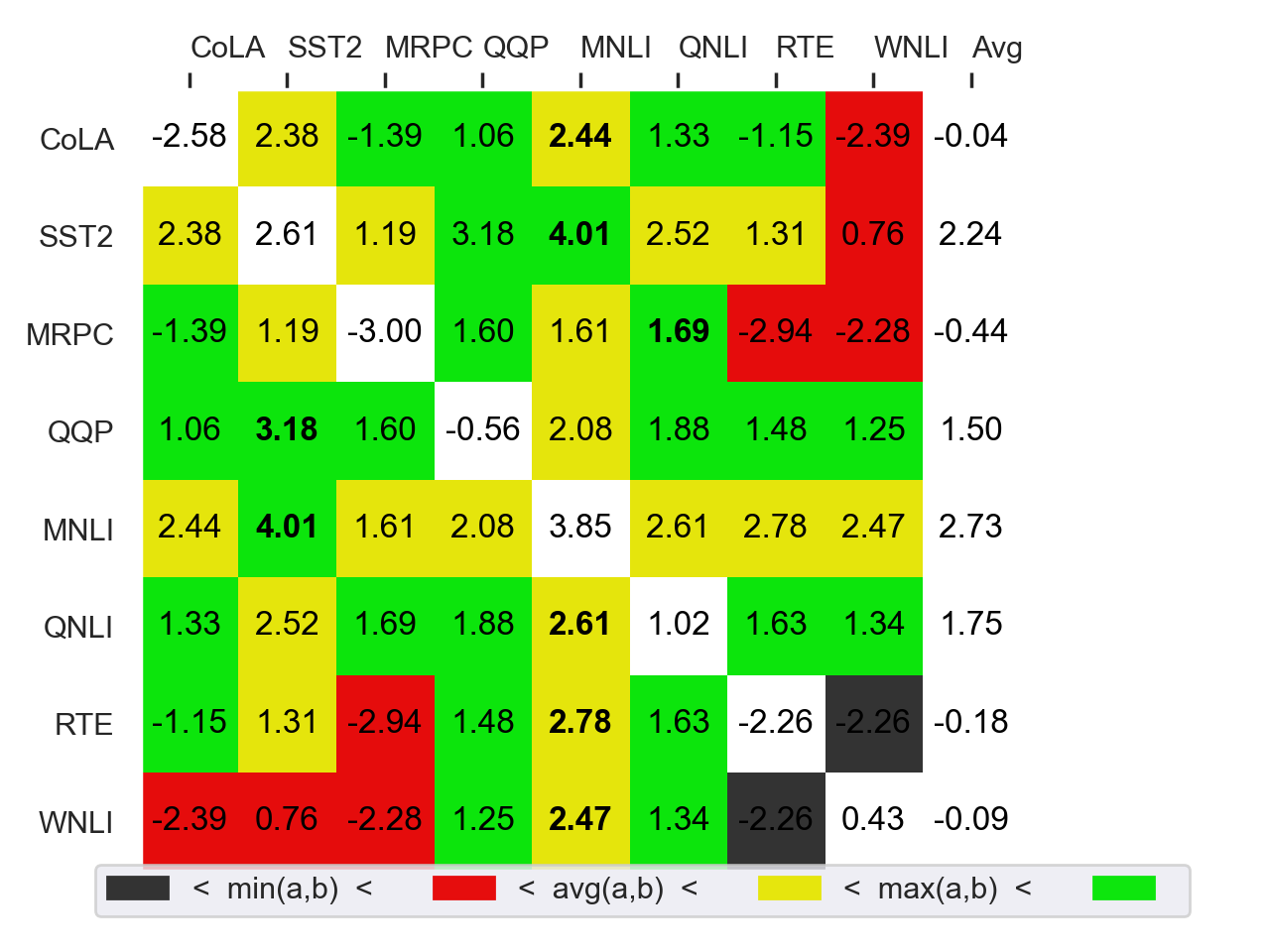}
\caption{Fusing two models is often better than the maximum intertraining achieved by either model. Average accuracy improvement over the pretrained on GLUE datasets. In the diagonal are intertraining, in each cell the result of fusing 2 models - finetuned on the column or row datasets (symmetrical). Colors compare fusing to intertraining with one of the models. Green represents fusing improves over both intertrainings, yellow over their mean, red improves over at least one, and Black is worse than both.
\label{fig:pairs}}
\end{figure}

We observe (Fig.~\ref{fig:pairs}) that careful choice of models to fuse, improves results dramatically. Specifically, fusing models which are also beneficial for intertraining yields best results.

Moreover, fusing pairs of models fares well in comparison to intertraining. In all but one pair, fusing outperforms the worst of the two models. In most cases, it outperforms the average result between them. Most importantly,
fusing is often even better than the maximum of the two models. Fusing MNLI and SST2 also reaches the highest results on our experiment.


\subsection{Sensitivity of Intertraining}\label{sec:decay}
All the experiments reported above has been trained without weight decay. Weight decay regularization factor is an addition to the loss that favors models with lower weight norms. 
AdamW with decay \citep{loshchilov2017decoupled} is considered a good choice for an optimizer in general and for finetuning in particular \footnote{For example, it is the default in HuggingFace \href{https://huggingface.co/docs/transformers/main_classes/trainer}{trainer} and \href{https://huggingface.co/course/chapter3/4?fw=pt}{course} and recommended by FastAI \cite{openai2018Optimizers}}. 

Regardless, we find intertraining with decay is not better than pretraining (Table~\ref{tab:decay_vs_no_decay}). Fusing, however, seems not to be affected as much and proves beneficial and robust even under such circumstances.
In initial trials with BERT \citep{devlin2018bert} which was pretrained with decay, showed less of this adverse effect. We deduce that intertraining should take into account the way pretraining was done.

We note that the results with weight decay are lower in general.

\begin{table}
\centering
\begin{tabular}{cccc}
                 & Pretrain     & Intertrain          & Fuse  \\
\midrule
T5 w/o decay     & 63.87        & \textbf{72.76}      & 68.12 \\
T5 with decay    & 61.6         & 61.7                & \textbf{65.1}  \\
\end{tabular}%
\caption{Results on general datasets with and without weight decay.\label{tab:decay_vs_no_decay}}
\end{table}

\subsection{Source Train Size}
\begin{figure}[tbp]
\includegraphics[width=8cm]{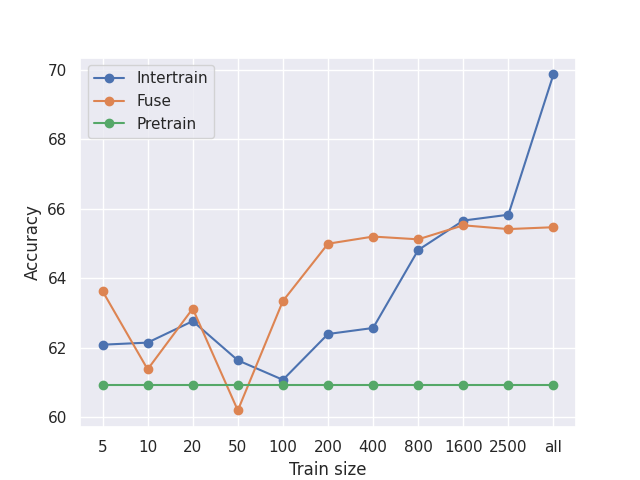}
\caption{Increasing source data results in better fused models. Accuracy on general target datasets by amount of source data for each model (log scale).}
\label{fig:source_no_decay_size}
\end{figure}

We consider the effect of source training size on fusing (See Fig.~\ref{fig:source_no_decay_size}). Evidently more data is better. Another, noisier trend is that fusing gains from smaller amounts of data than intertraining but its improvement also plateaus faster. This could be due to its use of data from several models.

\section{Related Work} \label{sec:back}
Since the rise of pretrained models various methods were proposed to improve them before tuning on the target task, such as further pretraining on the target task \citep{Gururangan2020DontSP} or learning to cluster it  \citep{shnarch2020cluster}. Those methods are applied to any base model and are hence complementary to ours. 

Intertraining was shown to be more beneficial than simple multitask learning \citep{Phang2018SentenceEO}. However, massively multitask learning was shown to be beneficial and to consistently do so when the number of tasks was increased \citep{Aghajanyan2021Muppet}. Our work does not  directly compete with massively multitask learning, as we only assume access to finetuned models, and not to the source data and training resources needed for such training. We do, however, build our intuition upon theirs (For more motivational findings see App~\ref{ap:sec:related_work}). Multi-task works show that when relying on many tasks, improvements are consistent. This may imply that there are features learnt for different tasks that may be shared. Moreover, they show that some families of tasks benefit from similar such features \citep{aribandi2021ext5}. We discuss fusing of similar and dissimilar tasks, finding fusing does not require a direct relation between the source and target tasks (see \S\ref{sec:results}). We do find such a requirement in intertraining.

Neural networks learn highly non-linear functions. Moreover, the dependency on each weight is non-linear. Therefore, the average of model weights does not compute the average function the networks encode. Indeed, research on models' similarity do not show models to be similar per weight \citep{Patel2021OnNI}. Yet, for the fusing by averaging weights to be beneficial, this should be true to some extent. Research does show similarity in other aspects such as in activations \citep{Raghu2017SVCCASV,Cohen2020SeparabilityAG}, in classification \citep{Hacohen2020LetsAT} or in their generalizations \citep{Choshen2021TheGT}. To sum, There is little theory to support why averaging model weights would be beneficial, and in the general case it probably is not. There is reason to believe that pretraining plays a crucial part in it \citep{neyshabur2020being}, and that the similar initialization allows for meaningful information to be gained by averaging. We leave it to future work.

An important distinction to make is between meta learning and transfer learning \citep{hospedales2020meta}. Transfer learning starts from a well performing model on one task and adapts it to, hoping it would be better in another task. Meta learning on the other hand, chooses a starting point that may not even perform any task, but that adapts well to new tasks. Our work takes pretraining and intermediate training which are transfer learning to the new domain of meta-learning. Where a fused model may not perform any of the tasks, it presumably starts at the point of least (Euclidean) distance from each model.  This point requires less finetuning than the initial one. In that sense, fusing models is somewhat reminiscence of one step of REPTILE meta learning \citep{Nichol2018reptile}. Rethinking of the starting point as a meta-learning problem also suggests that some errors introduced by averaging could still be fixed in the later training.

In parallel to our work \citet{matena2021merging,Wortsman2022ModelSA} followed some similar notions, in a transfer learning setting. In their works, they also propose to recycle finetuned models, specifically, they fuse models and then use the results without further finetuning. Their approaches however, consider a different setting. \citet{Wortsman2022ModelSA} focus on images and models finetuned on the same target task. \citet{matena2021merging} investigate whether, after finetuning on the target task, receiving another finetuned model could help. They also mostly consider the case where two models are being tuned (target task and another). An intriguing addition they make is approximating Fisher information to choose how much of each weight to take, following \citet{Kirkpatrick2017OvercomingCF}. Intuitively, given a dataset Fisher information indicates which weights contribute the most to the prediction. This approach is less suitable to our case where the base model should be general, and neither the source data of the fused models nor the target data is available.

\section{Future work}
Future work intends to delve further into the possibility of improving methods for fusing and understanding what characteristics of target tasks and models work well together. Furthermore, we expect to inspect fusing in a meta-learning scenario, where several iterations of fusing and finetuning occurs. Last, we may consider ways to fuse models of different sizes \citep{cotler2022analyzing}.

\clearpage
\bibliography{anthology,custom}
\bibliographystyle{acl_natbib}

\clearpage
\appendix

\section{How common is finetuning}\label{sec:tuning_freq}

Currently, finetuned models are shared mostly for reproducibility or inference on their target task and often fails to generalize to other domains \citep{Csordas2021TheDI, Ontanon2021MakingTS}. As it is, pretrained models are more important to the community.  This is exemplified by the number of downloads, where \href{https://huggingface.co/t5-small}{T5-Small} has 3M downloads at the month of writing, where its most popular finetuned version has 30K. Presumably, if finetuned models were useful as a starting point, they would have been more widely shared. 

We estimate finetuned models are indeed commonly created, supporting our assumption. A platform to share models exist and models are indeed shared there. The \href{https://huggingface.co/models}{HuggingFace} hub currently hosts 27K models, most of which are finetuned models, a handful are pretrained. However, at the current time, most tuned models are not uploaded to it. 

We validate that finetuning is indeed common. The downloads of pretrained models may hint to the number of finetunings performed, in the case of T5 those are millions a month. However, We estimate the amounts of unique tunings done in published research to give a broader picture. We arbitrarily chose 20 papers from EMNLP 2021, out of which 14 finetuned models on various tasks and none pretrained a model. In EMNLP alone there were above 800 papers published and in ACL 2021 top venues 3230 \footnote{Main papers in conferences, findings and CL as shown in \url{https://aclanthology.org/}} which gives an estimate of 560 and 2261 papers respectively that finetuned at least one model.

\section{Datasets used}\label{ap:sec:datasets}

All datasets could be downloaded from \href{https://huggingface.co/datasets/}{huggingface datasets}.
As we used groups of datasets we report here the full list of datasets they contain.

\paragraph{General}
GLUE:CoLA \citep{warstadt-etal-2019-neural}, SST2 \citep{socher-etal-2013-recursive}, MRPC \citep{dolan-brockett-2005-automatically}, QQP (\href{https://data.quora.com/First-Quora-Dataset-Release-Question-Pairs}{\texttt{data.quora.com/\allowbreak First-\allowbreak Quora-\allowbreak Dataset-\allowbreak Release-Question-Pairs}}), MNLI \citep{williams-etal-2018-broad}, QNLI \citealt{rajpurkar-etal-2016-squad}, RTE \citep{Dagan2005ThePR,BarHaim2006TheSP,Giampiccolo2007TheTP,Bentivogli2009TheSP}, WNLI \citep{Levesque2011TheWS}

SuperGLUE:BoolQ \citep{clark-etal-2019-boolq}, CB \citep{demarneffe:cb}, CoPA \citep{roemmele2011choice}, MULTIRC \citep{khashabi2018looking}, WIC \citep{pilehvar2018wic}, WSC \citep{levesque2012winograd}

\paragraph{NLI:} MNLI \citep{williams-etal-2018-broad}, QNLI \citealt{rajpurkar-etal-2016-squad}, RTE \citep{Dagan2005ThePR,BarHaim2006TheSP,Giampiccolo2007TheTP,Bentivogli2009TheSP}, WNLI \citep{Levesque2011TheWS}, ESNLI \citep{Camburu2018eSNLINL}, adversarial NLI \citep{nie-etal-2020-adversarial}.

\paragraph{TweetEval:}EmoInt \citep{MohammadB17starsem}, Emoji \citep{semeval2018task2}, Irony \citep{van-hee-etal-2018-semeval}, OffenseEval \citep{zampierietal2019}, HatEval \citep{basile-etal-2019-semeval}, Sentiment Analysis \citep{rosenthal-etal-2017-semeval}, Stance \citep{StanceSemEval2016}

As GLUE and SUPER-GLUE test sets are held out, whenever such datasets are used, we extracted 1K or 10\% of the training examples as test set, the smaller. We opted for different test sizes as some datasets are small, even smaller than 1K examples in total.
For MNLI we use the mismatched validation set as a test and the matched as a validation set.



\section{Standard Deviation}\label{ap:sec:std}
We present the standard deviation of experiments reported in the paper. The deviation given available models trained on one source task and tested on another are shown in Table~\ref{ap:tab:cross_std}.

\begin{table}[htbp]
\small
\begin{tabular}{ c c c c c | c }
 &  & General & NLI & Twitter & Average \\
\hline
& NLI & 1.42 & \textbf{0.64} & 2.76 & 1.61 \\
Intertain  & Twitter & 1.82 & 1.56 & 3.34 & 2.24 \\
& General & \textbf{1.02} & 2.83 & 2.66 & 2.17 \\
\hline
& NLI & 1.04 & 0.69 & \textbf{1.91} & \textbf{1.21} \\
Fuse  & Twitter & 1.98 & 1.78 & 3.04 & 2.27 \\
& General & 1.45 & 1.41 & 2.29 & 1.72 \\
\hline
Pretrain  &  & 3.50 & 1.67 & 5.75 & 3.64 \\
\end{tabular}
\caption{The standard deviation of both intertraining and fusing is better than finetuning on pretrained model. \label{ap:tab:cross_std}}
\end{table}

\section{Motivational Findings}\label{ap:sec:related_work}
We note two phenomena that give context to fusing. For fusing to make sense, the euclidean space in which model weights exist should have a strong connection to the loss. For example, if there are many local optima and each model picks another, then averaging would not result in a meaningful initialization.

Our approach to fusing was motivated by few work that supports the intuition that 
averaging could accumulate learnt information in some sense. Some works analyzed the loss spaces, noticing the Monotonic Linear Interpolation phenomena \citep{Goodfellow2015QualitativelyCN}, namely, the loss monotonically decreases when interpolating between a random initialization and a trained model. This was shown to be robust to changing most of the training parameters, but often fails or is less pronounced when adaptive optimizers are used \citep{Lucas2021AnalyzingML} and in other modern settings \citep{Frankle2020RevisitingC, Lucas2021OnML}. While we do not interpolate in such a way, and start from a pretrained model rather than random initialization, it at least supports the assumption that the euclidean space of model weights is meaningful and averaging vectors (finetuned models) within it, may lead to an advancement in the unknown general loss space of all textual tasks. 

Another studied phenomenon is the linear mode connectivity. Two models trained on the same task are connected if their linear combinations have similar loss. Mode connectivity is not a general trait of neural networks, but some features like deep networks and low loss contribute to its prevalence \citep{Entezari2021TheRO}. When there is mode connectivity, distributed weight averaging techniques should work \citep{Scaman2019OptimalCR}. \citet{Mirzadeh2021LinearMC} found linear mode connectivity between multitask and continual learning given the same initialization.
Last, \citet{Frankle2020LinearMC} suggests that due to the former, later in training, small changes to the weights are less harmful, if further tuned after them. We can consider the non-optimal weights found by fusing as adding noise.

\end{document}